\pdfoutput=1

\documentclass[11pt]{article}

\usepackage{acl}

\usepackage{times}
\usepackage{latexsym}
\usepackage{url}
\usepackage{amssymb}
\usepackage{amsfonts}
\usepackage{amsmath}
\usepackage[belowskip=-10pt,aboveskip=0pt,skip=10pt]{caption}
\usepackage{subcaption}
\usepackage{graphicx}
\usepackage{arydshln}
\usepackage{hyperref}
\usepackage{xcolor}
\usepackage{multirow}
\usepackage{tikz}
\usepackage{pgfplots}
\usepackage[export]{adjustbox}
\usepackage{marvosym}

\usepackage[T1]{fontenc}

\usepackage[utf8]{inputenc}

\usepackage{microtype}

\definecolor{butter}{rgb}{0.988,0.914,0.310}
\definecolor{chocolate}{rgb}{0.914,0.725,0.431}
\definecolor{chameleon}{rgb}{0.541,0.886,0.204}
\definecolor{skyblue}{rgb}{0.447,0.624,0.812}
\definecolor{plum}{rgb}{0.678,0.498,0.659}
\definecolor{scarletred}{rgb}{0.937,0.161,0.161}
\definecolor{myblue}{rgb}{0.192,0.510,0.729}
\definecolor{mygreen}{rgb}{0.173,0.627,0.173}
\definecolor{myorange}{rgb}{1.000,0.498,0.055}
\definecolor{myred}{rgb}{0.839,0.153,0.157}

\newcommand{\bs}[1]{\boldsymbol{#1}}

\title{Improving abstractive summarization with energy-based re-ranking}

\author{Diogo Pernes\textsuperscript{\Moon\Scorpio} \quad
        Afonso Mendes\textsuperscript{\Moon} \quad
        André F.~T. Martins\textsuperscript{\Neptune\Pluto\Saturn} \\
\textsuperscript{\Moon}Priberam~
\textsuperscript{\Scorpio}Universidade do Porto\\
\textsuperscript{\Neptune}Instituto de Telecomunicações~
\textsuperscript{\Pluto}LUMLIS (Lisbon ELLIS Unit), Instituto Superior T\'ecnico~
\textsuperscript{\Saturn}Unbabel\\
Lisbon, Portugal\\
\href{mailto:diogo.pernes@priberam.pt}{\tt diogo.pernes@priberam.pt},\\
\href{mailto:amm@priberam.pt}{\tt amm@priberam.pt}, \quad
\href{mailto:andre.t.martins@tecnico.ulisboa.pt}{\tt andre.t.martins@tecnico.ulisboa.pt}.
}

\begin{document}
\maketitle
\begin{abstract}
Current abstractive summarization systems present important weaknesses which prevent their deployment in real-world applications, such as the omission of relevant information and the generation of factual inconsistencies (also known as \emph{hallucinations}). At the same time, automatic evaluation metrics such as CTC scores \cite{deng-etal-2021-compression} have been recently proposed that exhibit a higher correlation with human judgments than traditional lexical-overlap metrics such as ROUGE. In this work, we intend to close the loop by leveraging the recent advances in summarization metrics to create \emph{quality-aware} abstractive summarizers. Namely, we propose an energy-based model that learns to re-rank summaries according to one or a combination of these metrics. We experiment using several metrics to train our energy-based re-ranker and show that it consistently improves the scores achieved by the predicted summaries. Nonetheless, human evaluation results show that the re-ranking approach should be used with care for highly abstractive summaries, as the available metrics are not yet sufficiently reliable for this purpose.
\end{abstract}

\section{Introduction}
\label{intro}

In recent years, abstractive methods have greatly benefited from the development and widespread availability of large-scale transformer-based language generative models \cite{Vaswani2017,lewis-etal-2020-bart,Raffel2020,Zhang2020}, which are capable of generating text with unprecedented fluency. Despite the recent progress, abstractive summarization systems still suffer from problems that hamper their deployment in real-world applications. Omitting the most relevant information from the source document is one of such problems. Additionally, factual inconsistencies (also known as \emph{hallucinations}) were estimated to be present in around 30\% of the summaries produced by abstractive systems on the CNN/DailyMail dataset \cite{kryscinski-etal-2019-neural}. This observation has motivated a considerable amount of research on strategies to mitigate the hallucination problem \cite{falke-etal-2019-ranking,cao-etal-2020-factual,zhao-etal-2020-reducing,zhu-etal-2021-enhancing}, but the improvements achieved so far are mild. This is partly due to the difficulty of evaluating the quality of summaries automatically, leading to the adoption of metrics that are often insufficient or even inappropriate. Despite its limitations, ROUGE \cite{lin-2004-rouge} is still the de facto evaluation metric for summarization, mostly due to its simplicity and interpretability. However, not only does it correlate poorly with human-assessed summary quality \cite{Kane2019}, but it is also unreliable whenever the reference summary contains hallucinations, which unfortunately is not an uncommon issue in widely adopted summarization datasets \cite{kryscinski-etal-2019-neural,maynez-etal-2020-faithfulness}. For these reasons, the development of more reliable evaluation metrics with a stronger correlation with human judgment is also an active area of research \cite{kryscinski-etal-2020-evaluating,scialom-etal-2021-questeval,deng-etal-2021-compression}.

In this work, we propose a new approach to abstractive summarization via an energy-based model. In contrast to previous approaches, which use reinforcement learning to train models to maximize ROUGE or BERT scores \cite{Paulus2018,li-etal-2019-deep}, our EBM is trained to re-rank the candidate summaries the same way that the chosen metric would rank them -- a much simpler problem which is computationally much more efficient. This way, we are distilling the metric, which presents as a by-product an additional advantage: a quality estimation system that can be used to assess the quality of the summaries on the fly without the need of reference summaries. It should be remarked that any reference-free metric, can be used at inference time for re-ranking candidates from any abstractive summarization system, hence improving the quality of the generated summaries. Our re-ranking model can therefore leverage the advantages of recently proposed evaluation metrics over traditional ones, which are essentially two-fold: i) being able to better capture high-level semantic concepts, and ii) in addition to the target summary, these metrics take into account the information present on the source document, which is crucial to detect hallucinations. We demonstrate the effectiveness of our approach on standard benchmark datasets for abstractive summarization (CNN/DailyMail, \citet{Hermann2015}, and XSum, \citet{narayan-etal-2018-dont}) and use a variety of summarization metrics as the target to train our model on, showing the versatility of the method. We also conduct a human evaluation experiment, in which we compare our re-ranking model trained to maximize recent transformer-based metrics that aim to measure factual consistency and relevance (CTC scores, \citet{deng-etal-2021-compression}). Our proposed model yields improvements over the usual beam search on a baseline model and demonstrates the ability to distill target metrics. However, the human evaluation results suggest that re-ranking according to these metrics, while competitive, may yield lower quality 
summaries than those obtained by state-of-the-art 
abstractive systems trained with augmented data
and contrastive learning.

The remainder of the paper is organized as follows: in Section~\ref{sec:related_work}, we discuss the related work; in Section~\ref{sec:abstractive_summarization}, we do a brief high-level description of neural abstractive summarization systems and how different candidate summaries can be generated from them; in Section~\ref{sec:methodology}, we describe our methodology in detail, as well as the summarization metrics that we shall use to train our re-ranking model; Section~\ref{sec:experiments} presents the experimental results of our model and baselines, which include both automatic and human evaluation; in Section~\ref{sec:limitations}, we discuss the limitations of our approach and point some directions for future work, and we conclude this work with some final remarks in Section~\ref{sec:conclusion}.

\section{Related work}
\label{sec:related_work}
In the context of natural language generation, the idea of re-ranking candidates has been studied extensively for neural machine translation \cite{shen-etal-2004-discriminative,mizumoto-matsumoto-2016-discriminative,ng-etal-2019-facebook,salazar-etal-2020-masked,fernandes2022quality}, but only seldom explored for abstractive summarization. Among the former, the approach by \newcite{bhattacharyya-etal-2021-energy} is the most similar to ours as they also resort to an energy-based model to re-rank the candidates. However, they do not apply their method to abstractive summarization and their training objective is different than the one we shall define for our model: at each training step, they sample a pair of candidates, and the model is trained so that the difference between the energies of the two candidates is at least as large as the difference of their BLEU scores \cite{papineni-etal-2002-bleu}. Thus, their approach only exploits the information of two candidates at each training step. 
Recently, improved learning objectives such as contrastive losses have been proposed to enhance the quality of the predicted summaries, especially their factual consistency. \newcite{tang-etal-2022-confit}, \newcite{cao-wang-2021-cliff}, and \newcite{Liu2021} used data augmentation to generate both factual consistent and inconsistent sentences and used these in a contrastive learning objective to regularize the transformer learned representations. In a different line of work, \newcite{cao-etal-2020-factual} and \newcite{zhao-etal-2020-reducing} trained separate models on the task of correcting factual inconsistencies in the predicted summaries. \newcite{zhu-etal-2021-enhancing} presented a model that learns to extract a knowledge graph from the source document and uses it to condition the decoding step. \newcite{goyal-durrett-2021-annotating} trained a model to detect non-factual tokens and used it to identify and discard these tokens from the training data of the summarizer. 
\newcite{aralikatte-etal-2021-focus} modified the output distribution of the model to put more focus on the vocabulary tokens that are similar to the attended input tokens. Despite being sensible ideas, these techniques mostly focus on redefining the training objective of the model and disregard the opportunity to improve the summary quality at inference time, either by redesigning the sampling algorithm or using re-ranking. In a somewhat similar direction to ours, a contemporary work \cite{liu-etal-2022-brio} proposes using a ranking objective as an additional term on the usual negative log-likelihood loss. Similar to us, \newcite{liu-liu-2021-simcls} and \newcite{ravaut-etal-2022-summareranker} propose to use a trained re-ranker in as post-generation step. The former use a contrastive objective to learn a re-ranker that mimics ROUGE scores. The latter  employs a mixture of experts to train a re-ranker on the combination of ROUGE, BERT and BART scores.

\section{Abstractive summarization systems}
\label{sec:abstractive_summarization}
A typical abstractive summarization model approximates the conditional distribution $p(y \mid x)$, of summaries $y$ given source documents $x$, and works auto-regressively, exploiting the chain rule of probability:
\begin{equation}
    p(y \mid x) = \prod_{i=1}^{l+1} p(y^{(i)} \mid x, y^{0:(i-1)}), 
\end{equation}
where $y^{(0)}$ is a start-of-sequence token, the following $y^{(1)}, \dots, y^{(l)}$ are the tokens in the summary, from the beginning to the end, and $y^{(l+1)}$ is an end-of-sequence token. Typically, the parameters of this model are estimated under the maximum likelihood criterion, by minimizing the negative log-likelihood loss for a training dataset $\{(x_i,y_i)\}_{i=1}^n$ containing source documents $x_i$ paired with the respective reference summaries $y_i$.

Usually, the decoding process aims at finding the most likely sequence $y^*$ for the given $x$, i.e.\ $y^* \triangleq \operatorname{arg}\max_y p(y \mid x)$. Since searching for the most likely sequence is intractable due to combinatorial explosion, mode-search heuristics like greedy decoding and beam search are used in practice. Even if one could find the optimal sequence, it is not guaranteed that this would be the best summary for the given document. A primary reason for this is that the distribution learned by the model is only an approximation of the true conditional distribution, and preserves some background knowledge acquired during the unsupervised pre-training of the underlying language model. This is responsible for the presence of additional information in the summary that was not in the source document, which is the most frequent form of hallucination in summarization \cite{maynez-etal-2020-faithfulness}. Another source of problems is the noise in the training datasets, which are often scrapped automatically from the web with little human supervision \cite{kryscinski-etal-2019-neural}.

In essence, finding the optimal training objective and decoding algorithm to obtain the best summary remains an open problem. We take a step in this direction by sampling a set of candidate summaries $\{\hat{y}_1, \hat{y}_2, \dots, \hat{y}_k\}$ and then using a re-ranking model to choose the best one. To ensure diverse candidates, we experiment with diverse beam search \cite{Vijayakumar2016}, a modification of traditional beam search including a term in the scoring function that penalizes for repetitions across different beams.

\section{Energy-based re-ranking}
\label{sec:methodology}

\subsection{Formulation}
Formally, a summarization metric is a function $\phi: \mathcal{X} \times \mathcal{Y}^2 \mapsto \mathbb{R}$ that takes as input the source document $x \in \mathcal{X}$, the human-written reference summary $y \in \mathcal{Y}$, and the generated summary $\hat{y} \in \mathcal{Y}$, and outputs a scalar, usually in the unit interval, measuring the quality of the generated summary. Without loss of generality, throughout this work we assume that higher values of the metric indicate a better summary (as evaluated by the metric). Then, for a given summarization metric $\phi$, our goal is to find a reference-free function $E: \mathcal{X} \times \mathcal{Y} \mapsto \mathbb{R}$ with parameters $\theta$ such that, for two candidate summaries $\hat{y}$ and $\hat{y}'$ for the same document $x$ with reference summary $y$, $E(x,\hat{y}; \theta) < E(x,\hat{y}'; \theta)$ if and only if $\phi(x,y,\hat{y}) > \phi(x,y,\hat{y}')$. In the spirit of energy-based models \cite{LeCun2006}, $E$ should assign \emph{low energy} wherever $p(y \mid x)$ is high and \emph{high energy} wherever $p(y \mid x)$ is low, but does not need to be normalized as a proper density. More precisely, $E$ should satisfy $p(y \mid x) \propto \exp(-E(x,y; \theta))$.  Under this perspective, at training time, $\phi$ works as a proxy for the true conditional distribution, which is unknown. At inference time, sampling summaries directly from the distribution defined by the energy-based model is a non-trivial task since this model is not defined auto-regressively \cite{Eikema2021}, unlike standard encoder-decoder models for summarization. Hence, we use its scores to re-rank candidate summaries previously obtained from a baseline summarization model.

\subsection{Training and inference}
We assume to have access to a training data set $\mathcal{D}=\{(x_i,y_i,\hat{\mathbf{y}}_i)\}_{i=1}^n$, where $x_i$ and $y_i$ are respectively the $i$-th source document and the corresponding reference summary and $\hat{\mathbf{y}}_i = \{\hat{y}_{i,1}, \hat{y}_{i,2}, \dots, \hat{y}_{i,k}\}$ is a set of (up to) $k$ candidate summaries sampled from a baseline summarization model, such as BART \cite{lewis-etal-2020-bart} or PEGASUS \cite{Zhang2020}. Several techniques have been proposed for training energy-based models that avoid the explicit computation of the partition function $Z(x;\theta) \triangleq \int_{\mathcal{Y}} \exp(-E(x,y; \theta))\, \mathrm{d}y$ and its gradient, which are usually intractable \cite{Song2021}. Here, given this data and the metric $\phi$, we adopt the ListMLE ranking loss \cite{Xia2008} as the training objective. Specifically, the model is trained to minimize:
\begin{equation}
\small
    \label{eq:loss}
    \mathcal{L}_{\phi}(\theta) \triangleq - \mathbb{E}_{(x,y,\hat{\mathbf{y}}) \sim \mathcal{D}} \log \prod_{i=1}^k \frac{\exp(-E(x,\hat{y}_i;\theta)/\tau)}{\sum_{j=i}^k \exp(-E(x,\hat{y}_j;\theta)/\tau)},
\end{equation}
where $\tau > 0$ is a temperature hyperparameter and the candidates $\hat{y}_1, \hat{y}_2, \dots, \hat{y}_k$ are sorted such that if $i < j$ then $\phi(x,y,\hat{y}_i) \geq \phi(x,y,\hat{y}_j)$.

To gain some intuition about this loss function, let us define: i) $\mathrm{r}_i$ as the random variable corresponding to the $i$-th ranked summary in a list of $k$ candidates $\hat{y}_1, \hat{y}_2, \dots, \hat{y}_k$ and ii) the probability that $\mathrm{r}_1$ takes the value $\hat{y}_1$ as:
\begin{equation}
    P(\mathrm{r}_1 = \hat{y}_1 \mid x) \triangleq \frac{\exp(-E(x,\hat{y}_1)/\tau)}{\sum_{j=1}^k \exp(-E(x,\hat{y}_j)/\tau)},
\end{equation}
where we have omitted the parameters $\theta$ for brevity. Assuming that the first $i-1$ candidates are ranked correctly, the probability that the $i$-th candidate is also ranked correctly is the probability that it is ranked first in the list $\hat{y}_i, \hat{y}_{i+1}, \dots, \hat{y}_k$, thus:
\begin{align}
    P(\mathrm{r}_i = \hat{y}_i \mid x,& \mathrm{r}_{1:(i-1)} = \hat{y}_{1:(i-1)}) = \nonumber \\ &= \frac{\exp(-E(x,\hat{y}_i)/\tau)}{\sum_{j=i}^k \exp(-E(x,\hat{y}_j)/\tau)}.
\end{align}
It then follows from the chain rule that the probability that all the $k$ candidates are ranked correctly is:
\begin{align}
    & P(\mathrm{r}_{1:k} = \hat{y}_{1:k} \mid x) = \nonumber\\
    &= \prod_{i=1}^k P(\mathrm{r}_i = \hat{y}_i \mid x, \mathrm{r}_{1:(i-1)} = \hat{y}_{1:(i-1)}) \nonumber \\
    &= \prod_{i=1}^k \frac{\exp(-E(x,\hat{y}_i)/\tau)}{\sum_{j=i}^k \exp(-E(x,\hat{y}_j)/\tau)}.
\end{align}
Hence, $P(\mathrm{r}_{1:k} \mid x)$ is a distribution over all the possible permutations of the $k$ candidates and the minimization of the loss $\mathcal{L}_{\phi}$ maximizes the likelihood of the correct permutation, i.e.\ of the permutation induced by ranking the candidates $\hat{y}_1, \dots, \hat{y}_k$ according to the metric $\phi(x,y,\cdot)$. At inference time, given an unsorted list $\hat{\mathbf{y}}$ of $k$ candidate summaries for the document $x$, we choose the candidate $\hat{y}^*$ that is the most likely to be the top-ranked:
\begin{equation}
    \hat{y}^* \triangleq \operatorname{arg} \max_{\hat{y} \in \hat{\mathbf{y}}} P(\mathrm{r}_1=\hat{y} \mid x)
    = \operatorname{arg} \min_{\hat{y} \in \hat{\mathbf{y}}} E(x,\hat{y}).
\end{equation}

Thus, our energy based-model aims at ranking a set of candidates the same way that the metric $\phi$ would rank them, but it does this without having access to the reference summary $y$. Therefore, this is a way to distill the information contained in the metric into a single and reference-free model that can rank summary hypotheses on the fly. 

\subsection{Adopted metrics}
\label{sec:adopted_metrics}

So far, the definition of summarization metric we have provided was generic, so now we focus on describing the particular metrics we have used to train our model. Summarization metrics can be divided into two groups: \emph{reference-dependent} and \emph{reference-free}, depending on whether $\phi$ actually needs the reference summary or not. In the latter case, $\phi(x,y,\hat{y}) \equiv \varphi(x,\hat{y}) \; \forall y$, for some function $\varphi$. Thus, reference-dependent metrics are mostly used to evaluate and compare summarization systems, whereas reference-free metrics can also be used to assess summary quality on the fly. Therefore, training our energy-based model using reference-dependent metrics provides an indirect way to use these metrics for the latter purpose as well.

Automatically assessing the quality of a summary is a non-trivial task since it depends on high-level concepts, such as factual consistency, relevance, coherence, and fluency \cite{Lloret2018}. These are loosely captured by classical metrics \cite{Kane2019,kryscinski-etal-2019-neural} such as ROUGE, which essentially measure the $n$-gram overlap between $\hat{y}$ and $y$. However, in recent years, the availability of powerful language representation models like BERT \cite{devlin-etal-2019-bert} permitted and motivated the development of several transformer-based automatic metrics.

There are a few metrics based on question generation (QG) and question answering (QA) models \cite{wang-etal-2020-asking,durmus-etal-2020-feqa}. Among these, QuestEval \cite{scialom-etal-2021-questeval} exhibits the strongest correlation with human judgment. This metric uses a QG model to generate questions from both the source document $x$ and the candidate summary $\hat{y}$ and a QA model to get the answers from both, which are then compared to produce a score in the unit interval. In addition to the QA and QG models, QuestEval uses an additional model to determine the importance weight of each question generated from $x$.    Although being reference-free, this metric is computationally expensive, so it is important to investigate whether our model can produce a similar ranking more efficiently.

Following a different paradigm, \newcite{deng-etal-2021-compression} proposed a set of metrics for natural language generation tasks, named CTC scores, which are based on the notion of \emph{information alignment}. They define the alignment of a document $a$ to a document $b$, denoted $\operatorname{align}(a \to b)$, as a vector with the same length as $a$ where the $i$-th position is a scalar in $[0, 1]$ representing the confidence that the information in the $i$-th token of $a$ is grounded in $b$. For summarization tasks, two alignment-based metrics are proposed, one for factual consistency and the other for relevance, both achieving state-of-the-art results in correlation with human judgment. A generated summary $\hat{y}$ is consistent with its source document $x$ if all the information in $\hat{y}$ is supported by $x$, hence the consistency score is:
\begin{equation}
    \textnormal{CTC}_\textnormal{consistency}(x,\hat{y}) \triangleq \operatorname{mean}(\operatorname{align}(\hat{y} \to x)). \label{eq:consistency}
\end{equation}
For relevance, the authors argue that, besides being consistent, $\hat{y}$ should contain as much information as possible from the reference summary $y$, so they define the relevance score as:
\begin{align}
    &\textnormal{CTC}_\textnormal{relevance}(x,y,\hat{y}) \triangleq \nonumber \\ 
    &\triangleq \operatorname{mean}(\operatorname{align}(\hat{y} \to x)) \times \operatorname{mean}(\operatorname{align}(y \to \hat{y})). \label{eq:relevance}
\end{align}
Clearly, both metrics produce a score in the unit interval, being consistency reference-free and relevance reference-dependent.

\section{Experiments}
\label{sec:experiments}

\subsection{Datasets}
We evaluate our model and the baselines in two benchmark datasets for abstractive summarization: CNN/DailyMail \cite{Hermann2015} and XSum \cite{narayan-etal-2018-dont}, both containing news articles paired with their respective reference summaries. In XSum, each summary consists of a single sentence, while in CNN/DailyMail it can consist of three sentences or more. XSum is also known to be more abstractive and to have more hallucinations than CNN/DailyMail \cite{narayan-etal-2018-dont,maynez-etal-2020-faithfulness}.

\subsection{Baselines}
A BART model \cite{lewis-etal-2020-bart} trained on the usual maximum likelihood objective is our baseline. Summaries are sampled from this model using the usual beam search. In addition, we also compare our model with the following state-of-the-art methods: BRIO, by \newcite{liu-etal-2022-brio}, which employs a ranking loss as an additional term on the training of the abstractive system; CLIFF, by \newcite{cao-wang-2021-cliff}, which uses data augmentation techniques and contrastive learning to enhance the factual consistency of the summaries; DAE, proposed by \newcite{goyal-durrett-2021-annotating}, which detects and discards non-factual tokens from the training data; FASum, by \newcite{zhu-etal-2021-enhancing}, which incorporates knowledge graphs also to enhance factual consistency; SummaReranker, by \newcite{ravaut-etal-2022-summareranker}, which employs a mixture of experts to train a re-ranker on the combination of various metrics. In Appendix~\ref{sec:appendix_ablation}, we also experiment training the re-ranking model with the max-margin objective proposed by \newcite{bhattacharyya-etal-2021-energy} for machine translation and we present the results obtained by using a perfect re-ranker for $\textnormal{CTC}_\textnormal{consistency}$ and QuestEval, which is feasible since these metrics are reference-free.

\subsection{Implementation details}
Our energy-based re-ranking model (EBR-ListMLE) consists of a BERT that receives as input a pair $(x,\hat{y})$, of source document $x$ and candidate summary $\hat{y}$, and outputs the corresponding energy score $E(x,\hat{y})$. Candidates are sampled using diverse beam search \cite{Vijayakumar2016} on a BART encoder-decoder fine-tuned on the respective summarization dataset. Further implementations details are provided in Appendix~\ref{sec:appendix_further_implementation_details}. For reproducibility purposes, our code and trained models are also publicly available\footnote{\url{https://github.com/Priberam/SummEBR}}. Regarding the baselines, we use the official source code and model checkpoints for CLIFF and DAE. The latter is only evaluated on the XSum dataset since there is no checkpoint available for CNN/DailyMail. For the same reason, BRIO is only evaluated on CNN/DailyMail. For FASum, we use the released predicted summaries directly since this is the only resource available.

\subsection{Metrics}
\label{sec:exp_metrics}
We train our model using the metrics discussed in section~\ref{sec:adopted_metrics} as the target metric $\phi$. Specifically, we experiment with ROUGE-L, QuestEval, $\textnormal{CTC}_\textnormal{relevance}$, and $\textnormal{CTC}_\textnormal{relevance} + \textnormal{CTC}_\textnormal{consistency}$. ROUGE scores, QuestEval and CTC scores each belong to a different evaluation paradigm and so it is interesting to investigate their effect on our re-ranking approach. It is important to point out that $\textnormal{CTC}_\textnormal{consistency}$ is a reference-free metric whose computational complexity is similar to that of our re-ranker, so it is pointless to train our model based on that metric alone. Instead, we report the results using this metric directly for re-ranking in Appendix~\ref{sec:appendix_ablation}. However, combining (i.e.\ summing) it with $\textnormal{CTC}_\textnormal{relevance}$ yields an interesting metric as it takes into account two fundamental attributes of a summary: factual consistency and relevance. QuestEval is also reference-free but it is much more computationally intensive as it requires a question generation and a question answering step. Thus, we train our model with this metric and report the computational times for comparison. For evaluation, in addition to the aforementioned metrics, we also report results for ROUGE-1, ROUGE-2, and FactCC \cite{kryscinski-etal-2020-evaluating}, which is a metric based on NLI scores.

\subsection{Automatic evaluation}
\label{sec:automatic_eval}
\subsubsection{Comparison with the baselines}
\label{sec:comparison_with_baselines}

\begin{table*}[t]
\centering
\resizebox{\textwidth}{!}{
\begin{tabular}{l|ccccccc|ccccccc}
                                 & \multicolumn{7}{c|}{CNN/DailyMail}                                    & \multicolumn{7}{c}{XSum}                                              \\
                                 & \texttt{R1} & \texttt{R2} & \texttt{RL} & \texttt{QE} & \texttt{Cons} & \texttt{Rel}  & \texttt{FCC} & \texttt{R1} & \texttt{R2} & \texttt{RL} & \texttt{QE} & \texttt{Cons} & \texttt{Rel} & \texttt{FCC} \\ \hline
BART                             & $43.64$ & $20.75$ &   $40.52$   &   $43.28$  &   $95.01$   &   $61.75$      &  $55.68$ & $42.67$ & $19.42$ &  $34.48$       &   $28.27$   &   $83.18$  &   $52.23$  &  $26.28$ \\ \hdashline
BRIO                             & $\bs{47.97^*}$  & $\bs{24.06^*}$   & $\bs{44.86^*}$   &  $43.49$    & $89.61$   & $60.75$   & $33.05$ &  $-$ & $-$ & $-$ & $-$ & $-$ & $-$ & $-$ \\ \hdashline[0.5pt/5pt]
CLIFF                            & $43.86$  &  $20.88$  &  $40.63$  &  $43.28$    &  $94.68$  & $60.38$  & $55.85$ &  $\underline{44.50}$ & $\bs{21.41}$ &    $\underline{36.41}$ &  $\underline{29.34}$ & $82.57$ &  $51.92$ &  $24.86$ \\ \hdashline[0.5pt/5pt]
DAE                             &  $-$  &  $-$ &  $-$       &  $-$  &  $-$  &  $-$ & $-$ & $37.61$ &   $14.19$  &  $28.84$  & $29.20$   &  $79.45$ & $51.05$     & $19.46$ \\ \hdashline[0.5pt/5pt]
FASum                            & $40.40$ & $17.68$ &  $37.26$ & $42.87$ & $94.30$  &  $57.91$  & $51.20$ &  $30.22$     &  $9.97$       & $23.69$  &  $24.35$ & $75.45$  &  $39.42$  & $26.96$ \\ \hdashline[0.5pt/5pt]
SummaReranker & $\underline{45.07}$ & $\underline{21.73}$ & $\underline{41.87}$ & $43.61$ & $95.07$ & $62.49$ & $54.50$ & $\bs{44.93}$ & $\underline{21.40}$ & $\bs{36.76}$ & $28.76$ & $83.00$ & $52.75$ & $26.27$ \\ \hdashline
EBR [\texttt{RL}]  & $44.90$ & $21.58$ &   $41.75$ & $43.60$ & $95.01$ & $62.16$    & $54.95$ & $43.63$ & $20.28$ &     $\underline{35.78}$ & $28.55$ & $84.47$ & $52.92$  & $\bs{27.21}$ \\ \hdashline[0.5pt/5pt]
EBR [\texttt{QE}] & $44.07$ & $21.13$ &   $40.94$ & $\bs{44.27^*}$ & $95.71$ & $62.48$ & $59.23$  &   $42.94$ & $19.42$ &  $34.62$ & $\bs{29.89}$ & $83.34$ & $52.50$ & $26.34$ \\ \hdashline[0.5pt/5pt]
EBR [\texttt{Rel}]            & $44.04$ & $20.98$ & $40.85$ & $43.78$ & $\underline{95.93}$ & $\bs{63.40}$ &    $\underline{60.28}$ &  $43.39$ & $19.75$ & $35.03$ & $28.60$ & $\underline{85.49}$ & $\bs{54.80}$ & $26.28$ \\ \hdashline[0.5pt/5pt]
EBR [\texttt{Cons}+\texttt{Rel}]          & $43.88$ & $20.87$ & $40.69$ & $\underline{43.79}$ & $\bs{96.15}$ & $\underline{63.32}$ & $\bs{61.67^*}$ & $43.28$ & $19.72$ &   $34.92$ & $28.66$ & $\bs{86.03^*}$ & $\underline{54.74}$ & $\underline{27.12}$
\end{tabular}}
\caption{Results of our models and baselines on each of the automatic evaluation metrics. Bold font indicates best result, and the second best results are underlined. A $^*$ mark indicates that the difference to the second best result is statistically significant (approximate permutation test at 95\%). In the re-ranking models, the metric in brackets indicates the target metric $\phi$ used to train the re-ranker. (\texttt{R1}: ROUGE-1, \texttt{R2}: ROUGE-2, \texttt{RL}: ROUGE-L, \texttt{QE}: QuestEval, \texttt{Cons}: $\textnormal{CTC}_\textnormal{consistency}$, \texttt{Rel}: $\textnormal{CTC}_\textnormal{relevance}$, \texttt{FCC}: FactCC)}
\label{tab:automatic_eval}
\end{table*}

\begin{table}[t]
\centering
\small{
\begin{tabular}{l|ccc}
     & EBR & $\textnormal{CTC}_\textnormal{consistency}$ & QuestEval \\ \hline
Time & $1$ & $1.83$                       & $114.98$  
\end{tabular}}
\caption{Relative computation times of the reference-free scorers when scoring $1000$ (document, summary) pairs from CNN/DailyMail. The absolute computation time for EBR was $23$s.}
\label{tab:timing}
\end{table}

The results obtained by our model and baselines are presented in Table~\ref{tab:automatic_eval}. We used $8$ candidates for the re-ranking models and beam search with $8$ beams for the baselines. The effect of using different number of candidates for re-ranking is studied in Appendix~\ref{sec:appendix_num_candidates}. It is noticeable that the best results for all the metrics are obtained by the EBR models, except for the ROUGE scores, where BRIO, CLIFF, and SummaReranker often outperform our models. SummaReranker is likely the strongest competitor with our models, achieving close-to-best ROUGE scores in both datasets and outperforming the BART baseline in most of the remaining metrics.  Surprisingly, DAE and FASum score below BART in the vast majority of the automatic metrics. Unfortunately, the authors of DAE do not provide results for any of these metrics. Regarding FASum, the authors do provide the ROUGE scores for their model but they evaluate factual consistency using a custom metric, for which they did not release the implementation.

Among the re-ranking models, the best result for a given metric is obtained when the model is trained to re-rank according to that metric, as expected. It is also interesting to observe that training for a given metric generally yields improvements in the remaining metrics as well. This might be an indication that the ranking model learns a useful measure of summary quality, rather than exploiting possible loopholes of the metrics. The best model overall is arguably EBR-ListMLE trained for $\textnormal{CTC}_\textnormal{consistency}+\textnormal{CTC}_\textnormal{relevance}$, achieving close to best results in all the metrics except ROUGE scores, which are known to correlate less strongly with human judgment.

We also compared the inference time of our model with the computation time of the two reference-free metrics, $\textnormal{CTC}_\textnormal{consistency}$ and QuestEval\footnote{We used an $80$-core CPU Intel Xeon Gold 5218R @ $2.10$GHz with $800$GB of RAM and a GPU NVIDIA A100 with $80$GB of memory.}. We performed this experiment by sampling $1000$ (document, summary) pairs from the test set of the CNN/DailyMail dataset and computing the scores one by one (i.e.\ without mini-batching) using our model and each of the metrics. The results are in Table~\ref{tab:timing}. The computation time of $\textnormal{CTC}_\textnormal{consistency}$ is comparable to, but larger than, that of our EBR, with the difference explained by the fact that the former is based on a RoBERTa-large model \cite{zhuang-etal-2021-robustly} and the latter uses BERT-base. As argued before and confirmed by these results, the computation of QuestEval takes two orders of magnitude longer, which motivates distilling this metric into an EBR.

\subsubsection{Cross-model experiments}
\label{sec:cross_model}

\begin{table*}[t]
\centering
\resizebox{\textwidth}{!}{
\begin{tabular}{l|ccccccc|ccccccc}
                                 & \multicolumn{7}{c|}{CNN/DailyMail}                                    & \multicolumn{7}{c}{XSum}                                              \\
                                 & \texttt{R1} & \texttt{R2} & \texttt{RL} & \texttt{QE} & \texttt{Cons} & \texttt{Rel}  & \texttt{FCC} & \texttt{R1} & \texttt{R2} & \texttt{RL} & \texttt{QE} & \texttt{Cons} & \texttt{Rel} & \texttt{FCC} \\ \hline
PEGASUS                             & $43.19$ & $20.64$ & $36.74$ & $41.22$ & $92.27$ & $59.09$ & $41.13$  & $46.64$ & $23.79$ & $38.53$ & $28.55$ & $82.02$ & $53.32$ & $24.10$ \\ \hdashline
EBR [\texttt{RL}]  & $\bs{44.35^*}$ & $\bs{21.37^*}$ & $\bs{37.66^*}$ & $41.60$ & $92.54$ & $59.50$ & $42.45$ & $46.74$ & $\bs{24.28^*}$ & $\bs{39.16^*}$ & $28.52$ & $82.01$ & $51.87$ & $\bs{26.04^*}$ \\ \hdashline[0.5pt/5pt]
EBR [\texttt{QE}] & $43.70$ & $21.04$ & $37.17$ & $\bs{42.28^*}$ & $93.30$ & $60.04$ & $45.31$ & $46.43$ & $23.58$ & $38.40$ & $\bs{29.82^*}$ & $82.72$ & $53.38$ & $22.94$  \\ \hdashline[0.5pt/5pt]
EBR [\texttt{Rel}]            &$43.51$ & $20.80$ & $36.80$ & $41.62$ & $93.38$ & $\bs{61.05^*}$ & $44.19$ & $\bs{46.92}$ & $23.70$ & $38.50$ & $28.78$ & $83.18$ & $\bs{55.33^*}$& $22.57$ \\ \hdashline[0.5pt/5pt]
EBR [\texttt{Cons}+\texttt{Rel}]          &$43.36$ & $20.76$ & $36.75$ & $41.74$ & $\bs{93.82^*}$ & $60.98$& $\bs{46.10^*}$  &$\bs{46.92}$ & $23.79$ & $38.61$ & $28.82$ & $\bs{83.82^*}$ & $55.26$ & $23.60$
\end{tabular}}
\caption{Results of the cross-model experiment in which EBRs trained with summaries from BART are tested on re-ranking summaries from PEGASUS. Bold font indicates best result. A $^*$ mark indicates that the difference to the result of PEGASUS is statistically significant (approximate permutation test at 95\%). In the re-ranking models, the metric in brackets indicates the target metric $\phi$ used to train the re-ranker. (\texttt{R1}: ROUGE-1, \texttt{R2}: ROUGE-2, \texttt{RL}: ROUGE-L, \texttt{QE}: QuestEval, \texttt{Cons}: $\textnormal{CTC}_\textnormal{consistency}$, \texttt{Rel}: $\textnormal{CTC}_\textnormal{relevance}$, \texttt{FCC}: FactCC)}
\label{tab:cross_model_results}
\end{table*}

An interesting question to investigate is whether our model is learning a general approximation of the target metric $\phi$, rather than just learning to recognize features that correlate with $\phi$ but are specific to the summarization system that generated the candidates. For this purpose, we experiment using a different abstractive summarizer to generate the test candidates than the one that was used to generate the training candidates. Specifically, we apply the same EBR models as in Section~\ref{sec:comparison_with_baselines}, which were trained using summaries sampled from BART, to re-rank summaries obtained from PEGASUS \cite{Zhang2020}. Like before, we obtain $8$ candidate summaries for each source document using beam search. In this experiment, our baseline is PEGASUS with no re-ranking. The results are in Table~\ref{tab:cross_model_results} and confirm that our EBR models have learned to mimic the respective metrics faithfully. The best score for each of the metrics is achieved by the EBR model that was trained for that metric. Moreover, when evaluated with different metrics, these models tend to surpass the PEGASUS baseline in the vast majority of the cases.

\subsection{Human evaluation}
\label{sec:human_eval}

\begin{table}[t]
\centering
\small{
\begin{tabular}{l|ccc|ccc}
                      & \multicolumn{3}{c|}{CNN/DailyMail}                                                                     & \multicolumn{3}{c}{XSum}                                                                              \\
\multicolumn{1}{c|}{} & \multicolumn{1}{c}{FC} & \multicolumn{1}{c}{R} & \multicolumn{1}{c|}{F} & \multicolumn{1}{c}{FC} & \multicolumn{1}{c}{R} & \multicolumn{1}{c}{F} \\ \hline
CLIFF is better       &  $.17$                                       &    $.33$                           &    $\bs{.33}$                          &    $\bs{.25}$                                     &  $\bs{.32}$                             &  $\bs{.27}$                           \\ \hdashline[0.5pt/5pt]
Tie                   &  $.65$                                       &    $.24$                           &    $.40$                          &  $.63$                                       &  $.63$                             &   $.68$                          \\ \hdashline[0.5pt/5pt]
BART is better        & $\bs{.18}$                                        &  $\bs{.43}$                             &     $.27$                         &  $.12$                                       &  $.05$                             &  $.05$                         \\ \hline
EBR is better         &  $\bs{.13}$                                       &  $\bs{.30}$                             & $\bs{.24}$                             &   $\bs{.15}$                                      &  $.12$                             &   $\bs{.30}$                          \\ \hdashline[0.5pt/5pt]
Tie                   &   $.80$                                      & $.52$                               &  $.58$                            &  $.72$                                       &   $.77$                            &  $.63$                           \\ \hdashline[0.5pt/5pt]
BART is better        &  $.07$                                       &  $.18$                             & $.18$                             & $.13$                                         &  $.12$                             &  $.07$                         \\ \hline
EBR is better         &  $.12$                                       &  $\bs{.45}$                             &  $\bs{.32}$                            & $.10$                                       &   $.08$                            &    $.07$                         \\ \hdashline[0.5pt/5pt]
Tie                   &  $.68$                                       &  $.20$                             & $.42$                             & $.63$                                        &   $.63$                            &   $.88$                          \\ \hdashline[0.5pt/5pt]
CLIFF is better        & $\bs{.20}$                                        &   $.35$                            &   $.27$                           & $\bs{.27}$                                        &     $\bs{.28}$                          &   $\bs{.08}$                        \\  \hline
Agreement              &    $.50$                                     &     $.63$                          &  $.54$                            &   $.56$                                      & $.58$                              &   $.87$    \\ \hdashline[0.5pt/5pt]
Strong disag.              &    $.01$                                     &   $.11$                            & $.08$                             &  $.01$                                       &  $.00$                             & $.00$
\end{tabular}}
\caption{Proportion of times that each model was considered the best for the human judges in each pairwise comparison according to each criteria (FC: factual consistency, R: relevance, F: fluency). Rows “Agreement” and “Strong disag.” show, respectively, the proportion of times that the two judges agreed and chose opposite options on the pairwise comparisons.}
\label{tab:human_eval}
\end{table}

\begin{figure}
\includegraphics[width=7cm]{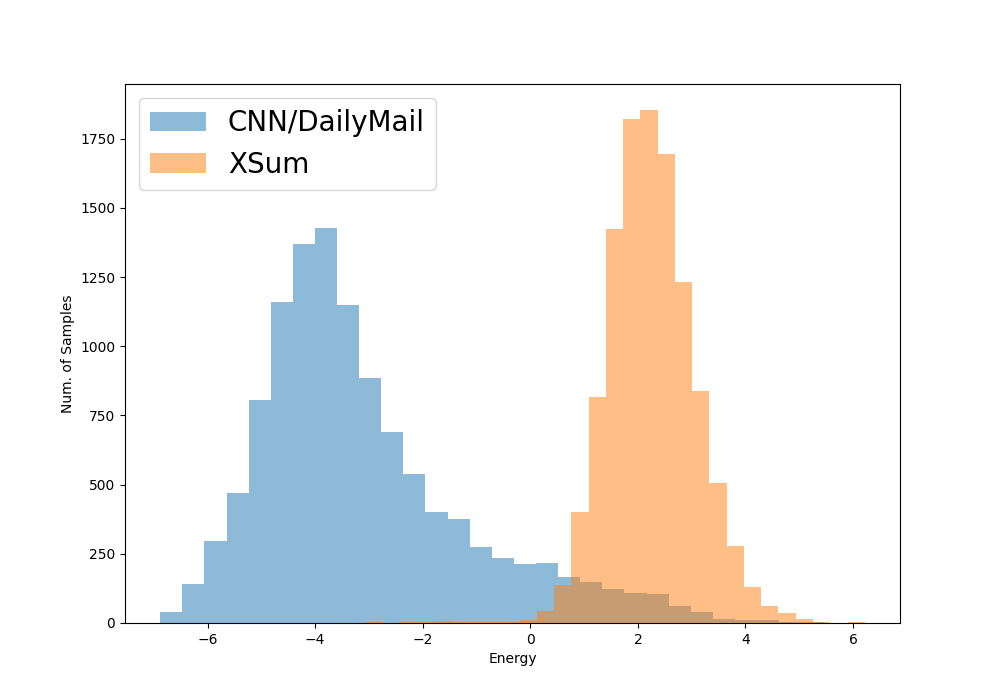}
\caption{Energy histogram of the candidate summaries chosen by the EBR model on CNN/DailyMail and XSum.}
\label{fig:energy_histogram}
\end{figure}

Even though the results on automatic evaluation are promising, directly optimizing a metric is risky as none of these metrics correlate perfectly with human judgment. For this reason, it is crucial to conduct human evaluation. Specifically, we asked the judges to do pairwise comparisons between the summaries generated by three models: BART, CLIFF, which was the strongest published baseline at the time we conducted this study, and our EBR trained for $\textnormal{CTC}_\textnormal{consistency}+\textnormal{CTC}_\textnormal{relevance}$ and re-ranking candidates from BART. We chose these metrics for the EBR since they exhibit stronger correlation with human judgment than the remaining \cite{deng-etal-2021-compression} and explicitly account for two key attributes of a summary: factual consistency and relevance. For each source document, we presented three pairs of summaries consecutively, which correspond to all the pairwise combinations of the summaries generated by the three systems. Then, we asked the judges to rank the summaries in each pair according to three criteria: factual consistency, relevance, and fluency. For each criterion, the judges had to evaluate whether the first summary was better than, tied with, or worse than the second summary. The names of the systems that generated each summary were not shown to the judges and the order at which summaries were presented was randomized. We randomly sampled $30$ source documents from the test set of CNN/DailyMail and another $30$ from the test set of XSum, so each judge was asked to compare $180$ pairs of summaries. A screenshot and description of the user interface of the evaluation form is provided in Appendix~\ref{sec:appendix_eval_form}. We recruited two judges for this task, who are specialists in linguistics. The results are presented in Table~\ref{tab:human_eval}. The first observation is that our EBR model succeeds at improving the quality of the candidates sampled from BART on the CNN/DailyMail dataset in all the three criteria. On XSum, the improvements are marginal or even absent, except on the fluency dimension. The EBR model itself has lower confidence on the predictions made on the XSum dataset: as shown in Figure~\ref{fig:energy_histogram}, the EBR model generally assigns higher energy to the XSum summaries than to the CNN/DailyMail summaries. The fact that our model improves fluency, which it was not trained for, may indicate that there is an implicit bias in our model and/or in the the target metrics ($\textnormal{CTC}_\textnormal{consistency}$ and $\textnormal{CTC}_\textnormal{relevance}$) towards more fluent summaries. Surprisingly, the comparison of our 
model with CLIFF contradicts the results 
of the automatic evaluation (Table~\ref{tab:automatic_eval}), especially 
on the XSum dataset. Three reasons could explain this phenomenon: i) the small number of documents used for human evaluation when compared to the size of the whole test set, ii) the EBR failing to re-rank the candidates according to the target metrics on these documents, and iii) limitations of the metrics themselves. In order to investigate which is true, we computed the actual values of $\textnormal{CTC}_\textnormal{consistency}$ and $\textnormal{CTC}_\textnormal{relevance}$ on the examples from XSum used for human evaluation. Regarding $\textnormal{CTC}_\textnormal{consistency}$, the summaries of EBR achieve a better score than those of CLIFF in $22$ cases (out of $30$), with an average score of $83.9\%$ vs.\ $80.2\%$ for CLIFF. For $\textnormal{CTC}_\textnormal{relevance}$, EBR wins against CLIFF in $20$ cases, with average scores of $54.3\%$ and $49.9\%$, respectively. We have also inspected the particular examples (shown in Appendix~\ref{sec:appendix_factual_inconsistencies}) where the judges agreed that CLIFF summary was better than the EBR summary on the factual consistency dimension. This happened only in three cases, but in all of them the EBR summary has obvious hallucinations and the CLIFF summary does not. Nonetheless, in two of them, the $\textnormal{CTC}_\textnormal{consistency}$ scores of the EBR summaries are larger than those of the CLIFF summaries, which confirms the flaws of the metric.

\section{Limitations and future work}

\label{sec:limitations}
Despite the improvements attained by our EBR model, its applicability is fundamentally dependent on the availability of reliable automatic evaluation metrics. Unfortunately, the correlation of these metrics with human judgment is still imperfect, especially for highly abstractive summaries. In addition, transformer-based metrics are currently only available for English. Finally, their backbone models are trained on news data, which hampers the reliability of these metrics in other domains. It is, therefore, crucial to continue the pursuit for more reliable metrics and to extend them to more languages and domains. 

\section{Conclusion}
\label{sec:conclusion}
We proposed an energy-based re-ranking model that can be trained to rank candidate summaries according to a pre-specified metric, leveraging the recent advancements in automatic summarization metrics to enhance the quality of the generated summaries. The experiments show that the proposed re-ranking model succeeds at distilling the target metrics, consistently improving the scores of the generated summaries. However, these improvements not always agree with the human evaluation, especially in the more abstractive setting (XSum), due to flaws of the adopted target metrics (CTC scores). Nonetheless, the proposed approach is flexible in the sense that we can train it with any target metric and apply it in conjunction with virtually any abstractive summarization system.

\section*{Acknowledgments}
This work is supported by the EU H2020 SELMA project (grant agreement No.\ 957017).

\bibliography{anthology,custom}
\bibliographystyle{acl_natbib}

\appendix

\section{Further implementation details}
\label{sec:appendix_further_implementation_details}

\subsection{Hyperparameters}
\label{sec:appendix_hyperparameters}
To generate the training data for the re-ranking model, we sample $8$ candidate summaries for each source document using diverse beam search with a diversity weight of $0.8$. The candidates are then ranked according to the desired metric $\phi$ and the BERT model is fine-tuned on this data for up to $4$ epochs, with a batch size of $24$, and using the Adam optimizer \cite{Kingma2014} with a learning rate of $5 \times 10^{-5}$. We use $\tau=1$ (equation~\eqref{eq:loss}) in all experiments. We keep the model that achieves the highest normalized discounted cumulative gain in a validation set. To generate the candidates at inference time, we set the diversity weight to zero since results in a separate validation set showed that this option yields the best results in most cases (see Appendix~\ref{sec:appendix_diversity_weight}). The models are implemented using the HuggingFace library on top of PyTorch. We also use HuggingFace publicly available checkpoints for the BART summarizers (\texttt{facebook/bart-large-cnn} and \texttt{facebook/bart-large-xsum}) and for BERT (\texttt{bert-base-uncased}).

\subsection{Choice of the diversity weight}
\label{sec:appendix_diversity_weight}

\begin{table*}[t!]
\centering
\small{
\begin{tabular}{lc|cccc}
     & Diversity weight       & \texttt{RL}      & \texttt{QE}      & \texttt{Cons}    & \texttt{Rel}     \\ \hline
EBR-ListMLE [\texttt{RL}] & \multirow{2}{*}{$0$}   & $43.50$ & $\bs{43.57}$ & $\bs{95.32}$ & $\bs{63.28}$  \\
Oracle [\texttt{RL}]      &                        & $46.95$ & $43.37$ & $95.26$ & $64.54$  \\ \hdashline[0.5pt/5pt]
EBR-ListMLE [\texttt{RL}] & \multirow{2}{*}{$0.2$} & $\bs{44.96}$ & $42.83$ & $90.61$ & $59.54$  \\
Oracle [\texttt{RL}]      &                        & $49.89$ & $42.48$ & $90.72$ & $61.39$  \\ \hdashline[0.5pt/5pt]
EBR-ListMLE [\texttt{RL}] & \multirow{2}{*}{$0.5$} & $44.98$ & $42.83$ & $90.47$ & $59.53$  \\
Oracle [\texttt{RL}]      &                        & $50.58$ & $42.44$ & $90.71$ & $61.61$  \\ \hdashline[0.5pt/5pt]
EBR-ListMLE [\texttt{RL}] & \multirow{2}{*}{$0.8$} & $44.92$ & $42.81$ & $90.32$ & $59.42$  \\
Oracle [\texttt{RL}]      &                        & $50.72$ & $42.38$ & $90.59$ & $61.65$  \\ \hline
EBR-ListMLE [\texttt{QE}] & \multirow{2}{*}{$0$}   & $42.59$ & $\bs{44.17}$ & $\bs{95.93}$ & $\bs{63.59}$  \\
Oracle [\texttt{QE}]      &                        & $42.55$ & $45.72$ & $95.72$ & $63.19$  \\ \hdashline[0.5pt/5pt]
EBR-ListMLE [\texttt{QE}] & \multirow{2}{*}{$0.2$} & $44.01$ & $43.92$ & $92.57$ & $60.82$  \\
Oracle [\texttt{QE}]      &                        & $43.80$ & $45.60$ & $91.84$ & $59.98$  \\ \hdashline[0.5pt/5pt]
EBR-ListMLE [\texttt{QE}] & \multirow{2}{*}{$0.5$} & $\bs{44.08}$ & $44.08$ & $92.69$ & $60.97$  \\
Oracle [\texttt{QE}]      &                        & $43.92$ & $45.84$ & $91.87$ & $60.15$  \\ \hdashline[0.5pt/5pt]
EBR-ListMLE [\texttt{QE}] & \multirow{2}{*}{$0.8$} & $43.95$ & $44.09$ & $92.70$ & $60.87$  \\
Oracle [\texttt{QE}]      &                        & $43.74$ & $45.88$ & $91.81$ & $60.00$  \\ \hline
EBR-ListMLE [\texttt{Rel}] & \multirow{2}{*}{$0$}   & $42.67$ & $\bs{43.70}$ & $\bs{96.11}$ & $\bs{64.53}$  \\
Oracle [\texttt{Rel}]      &                        & $44.32$ & $43.52$ & $96.24$ & $66.40$  \\ \hdashline[0.5pt/5pt]
EBR-ListMLE [\texttt{Rel}] & \multirow{2}{*}{$0.2$} & $43.83$ & $43.24$ & $93.77$ & $62.26$  \\
Oracle [\texttt{Rel}]      &                        & $46.04$ & $42.87$ & $93.56$ & $64.52$  \\ \hdashline[0.5pt/5pt]
EBR-ListMLE [\texttt{Rel}] & \multirow{2}{*}{$0.5$} & $\bs{43.87}$ & $43.32$ & $94.03$ & $62.51$  \\
Oracle [\texttt{Rel}]      &                        & $46.40$ & $42.92$ & $93.72$ & $65.10$  \\ \hdashline[0.5pt/5pt]
EBR-ListMLE [\texttt{Rel}] & \multirow{2}{*}{$0.8$} & $43.79$ & $43.29$ & $94.06$ & $62.47$ \\
Oracle [\texttt{Rel}]      &                        & $46.40$ & $42.82$ & $93.69$ & $65.18$  \\ \hline
EBR-ListMLE [\texttt{Cons}+\texttt{Rel}] & \multirow{2}{*}{$0$}   & $42.49$ & $\bs{43.69}$ & $\bs{96.35}$ & $\bs{64.45}$ \\
Oracle [\texttt{Cons}+\texttt{Rel}]      &                        & $44.09$ & $43.57$ & $96.56$ & $66.27$ \\ \hdashline[0.5pt/5pt]
EBR-ListMLE [\texttt{Cons}+\texttt{Rel}] & \multirow{2}{*}{$0.2$} & $\bs{43.62}$ & $43.24$ & $94.21$ & $62.25$ \\
Oracle [\texttt{Cons}+\texttt{Rel}]      &                        & $45.30$ & $43.00$ & $94.42$ & $64.20$ \\ \hdashline[0.5pt/5pt]
EBR-ListMLE [\texttt{Cons}+\texttt{Rel}] & \multirow{2}{*}{$0.5$} & $43.50$ & $43.24$ & $94.52$ & $62.44$ \\
Oracle [\texttt{Cons}+\texttt{Rel}]      &                        & $45.56$ & $43.09$ & $94.67$ & $64.74$ \\ \hdashline[0.5pt/5pt]
EBR-ListMLE [\texttt{Cons}+\texttt{Rel}] & \multirow{2}{*}{$0.8$} & $43.43$ & $43.21$ & $94.56$ & $62.46$ \\
Oracle [\texttt{Cons}+\texttt{Rel}]      &                        & $45.42$ & $43.02$ & $94.70$ & $64.79$ \\ 
\end{tabular}}
\caption{Results (in \%) for different diversity weights in a held-out validation set of CNN/DailyMail. (\texttt{RL}: ROUGE-L, \texttt{QE}: QuestEval, \texttt{Cons}: $\textnormal{CTC}_\textnormal{consistency}$, \texttt{Rel}: $\textnormal{CTC}_\textnormal{relevance}$)}
\label{tab:diversity_weight}
\end{table*}

Although we have used diverse beam search to generate the candidate summaries for training, we decided to stick to \textit{vanilla} beam search for testing. This choice was made based on the results presented in Table~\ref{tab:diversity_weight}. For this experiment, we have used a held-out development set from the validation set of CNN/DailyMail and we registered the results achieved by our EBR model and by an oracle re-ranker with diversity weights ranging from $0$ to $0.8$. According to all the metrics except ROUGE-L, setting the diversity weight to a positive value has a negative effect on the quality of the generated hypotheses since even an oracle re-ranker would have better results when the diversity weight is zero. Thus, we decided to set it at this value for the subsequent experiments with the test set.

\section{Ablation study}
\label{sec:appendix_ablation}
We now study the effect of training our EBR model using the max-margin loss proposed by \citet{bhattacharyya-etal-2021-energy} for machine translation. In addition, we also compare our models with perfect re-rankers for the two reference-free metrics: QuestEval and $\textnormal{CTC}_\textnormal{consistency}$. The results are in Table~\ref{tab:ablation}, where we also reproduce the results from our models presented in Table~\ref{tab:automatic_eval} for easier analysis. The comparison between the max-margin loss (EBR-MM) and ListMLE (EBR-ListMLE) shows that the latter tends to perform slightly better, although in the majority of the cases the difference is not statistically significant. It should also be remarked that re-ranking with the $\textnormal{CTC}_\textnormal{consistency}$ metric directly (Perfect Re-Rank [\texttt{Cons}]) yields competitive results too: it is the best on this metric in both datasets and it is close to the best model on $\textnormal{CTC}_\textnormal{relevance}$ in XSum. Re-ranking with QuestEval (Perfect Re-Rank [\texttt{QE}]) generally produces inferior results and, as shown previously in Table~\ref{tab:timing}, has the additional inconvenience of being much slower.

\begin{table*}[t]
\centering
\resizebox{\textwidth}{!}{
\begin{tabular}{l|ccccccc|ccccccc}
                                 & \multicolumn{7}{c|}{CNN/DailyMail}                                    & \multicolumn{7}{c}{XSum}                                              \\
                                 & \texttt{R1} & \texttt{R2} & \texttt{RL} & \texttt{QE} & \texttt{Cons} & \texttt{Rel}  & \texttt{FCC} & \texttt{R1} & \texttt{R2} & \texttt{RL} & \texttt{QE} & \texttt{Cons} & \texttt{Rel} & \texttt{FCC} \\ \hline
Perfect Re-Rank [\texttt{QE}]  &  $43.91$ & $20.99$ & $40.76$ & $\bs{45.74^*}$ & $95.46$ & $62.06$ &      $58.26$  & $42.81$ & $19.33$ & $34.53$ & $\bs{32.28^*}$ & $83.31$ & $52.57$ &  $26.58$ \\ \hdashline[0.5pt/5pt]
Perfect Re-Rank [\texttt{Cons}]  &  $43.43$ & $20.59$ & $40.29$ & $43.68$ & $\bs{96.69^*}$ & $62.60$ & $\underline{61.36}$  & $43.02$ & $19.58$ & $34.76$ & $28.70$ & $\bs{87.64^*}$ & $54.33$ &  $\bs{27.61^*}$ \\ \hdashline
EBR-MM [\texttt{RL}]  & $\underline{44.49}$ & $\underline{21.35}$ &   $\underline{41.32}$ & $43.72$ & $95.23$ & $62.22$   & $56.89$ &    $\bs{43.86}$ & $\bs{20.30}$ &     $\underline{35.72}$ & $28.68$ & $83.32$ & $52.85$    & $25.98$ \\ \hdashline[0.5pt/5pt]
EBR-MM [\texttt{QE}]  &  $44.07$ & $21.13$ &     $40.93$ & $44.22$ & $95.70$ & $62.54$     & $59.49$ & $42.85$ & $19.42$ &    $34.54$ & $29.63$ & $83.37$ & $52.58$    &  $25.86$ \\ \hdashline[0.5pt/5pt]
EBR-MM [\texttt{Rel}]  & $43.92$ & $20.87$ &   $40.72$ & $43.79$ & $95.78$ & $63.20$   & $59.84$ &   $43.44$ & $19.83$ &   $35.03$ & $28.79$ & $84.82$ & $54.54$ & $25.67$ \\ \hdashline[0.5pt/5pt]
EBR-MM [\texttt{Cons}+\texttt{Rel}]  & $43.75$ & $20.78$ &    $40.56$ & $43.78$ & $95.98$ & $63.10$ & $60.75$ & $43.31$ & $19.76$ &   $34.95$ & $28.83$ & $85.42$ & $54.50$    & $26.38$ \\ \hdashline
EBR-ListMLE [\texttt{RL}]  & $\bs{44.90}$ & $\bs{21.58}$ &   $\bs{41.75^*}$ & $43.60$ & $95.01$ & $62.16$    & $54.95$ & $\underline{43.63}$ & $\underline{20.28}$ &     $\bs{35.78}$ & $28.55$ & $84.47$ & $52.92$  & $\underline{27.21}$ \\ \hdashline[0.5pt/5pt]
EBR-ListMLE [\texttt{QE}] & $44.07$ & $21.13$ &   $40.94$ & $\underline{44.27}$ & $95.71$ & $62.48$ & $59.23$  &   $42.94$ & $19.42$ &  $34.62$ & $\underline{29.89}$ & $83.34$ & $52.50$ & $26.34$ \\ \hdashline[0.5pt/5pt]
EBR-ListMLE [\texttt{Rel}]            & $44.04$ & $20.98$ & $40.85$ & $43.78$ & $95.93$ & $\bs{63.40}$ &    $60.28$ &  $43.39$ & $19.75$ & $35.03$ & $28.60$ & $85.49$ & $\bs{54.80}$ & $26.28$ \\ \hdashline[0.5pt/5pt]
EBR-ListMLE [\texttt{Cons}+\texttt{Rel}]          & $43.88$ & $20.87$ & $40.69$ & $43.79$ & $\underline{96.15}$ & $\underline{63.32}$ & $\bs{61.67^*}$ & $43.28$ & $19.72$ &   $34.92$ & $28.66$ & $\underline{86.03}$ & $\underline{54.74}$ & $27.12$
\end{tabular}}
\caption{Results of our models (EBR-ListMLE) and baselines on each of the automatic evaluation metrics. Bold font indicates best result, and the second best results are underlined. A $^*$ mark indicates that the difference to the second best result is statistically significant (approximate permutation test at 95\%). The metric in brackets indicates the target metric $\phi$ used to train the re-ranker. (\texttt{R1}: ROUGE-1, \texttt{R2}: ROUGE-2, \texttt{RL}: ROUGE-L, \texttt{QE}: QuestEval, \texttt{Cons}: $\textnormal{CTC}_\textnormal{consistency}$, \texttt{Rel}: $\textnormal{CTC}_\textnormal{relevance}$, \texttt{FCC}: FactCC)}
\label{tab:ablation}
\end{table*}

\section{Effect of varying the number of candidates}
\label{sec:appendix_num_candidates}

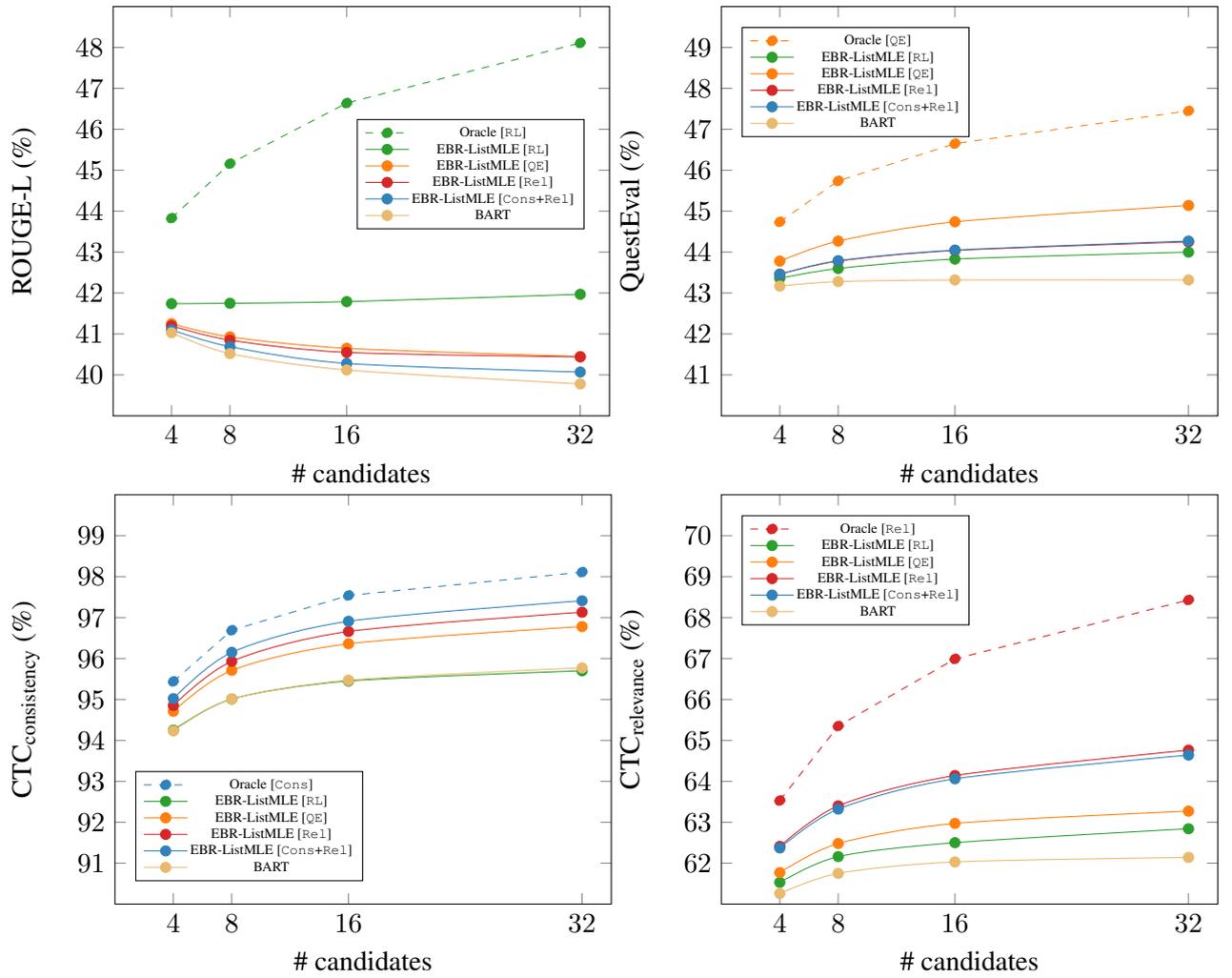
\begin{figure*}
\centering
\begin{subfigure}[b]{0.35\textwidth}
\centering
\begin{tikzpicture}
    \begin{axis}[
        xlabel=\# candidates,
        ylabel=ROUGE-L (\%),
        xmin=0, xmax=34,
        ymin=39, ymax=49,
        xtick={4,8,16,32},
        ytick={40,41,42,43,44,45,46,47,48},
        legend style={nodes={scale=0.5, transform shape}, at={(0.95,0.725)}}
        ]
    \addplot[dashed,mark=*,mygreen] plot coordinates {
        (4,43.83)
        (8,45.16)
        (16,46.64)
        (32,48.11)
    };
    \addlegendentry{Oracle [\texttt{RL}]}
    \addplot[smooth,mark=*,mygreen] plot coordinates {
        (4,41.74)
        (8,41.75)
        (16,41.79)
        (32,41.97)
     };
    \addlegendentry{EBR-ListMLE [\texttt{RL}]}
    \addplot[smooth,mark=*,myorange] plot coordinates {
        (4,41.25)
        (8,40.93)
        (16,40.65)
        (32,40.45)
    };
    \addlegendentry{EBR-ListMLE [\texttt{QE}]}
    \addplot[smooth,mark=*,myred] plot coordinates {
        (4,41.20)
        (8,40.85)
        (16,40.55)
        (32,40.44)
    };
    \addlegendentry{EBR-ListMLE [\texttt{Rel}]}
    \addplot[smooth,mark=*,myblue] plot coordinates {
        (4,41.10)
        (8,40.69)
        (16,40.28)
        (32,40.07)
    };
    \addlegendentry{EBR-ListMLE [\texttt{Cons}+\texttt{Rel}]}
    \addplot[smooth,mark=*,chocolate] plot coordinates {
        (4,41.03)
        (8,40.52)
        (16,40.12)
        (32,39.78)
    };
    \addlegendentry{BART}
    \end{axis}
\end{tikzpicture}
\end{subfigure}
\;\hspace{2.5cm}
\begin{subfigure}[b]{0.35\textwidth}
\centering
\begin{tikzpicture}
    \begin{axis}[
        xlabel=\# candidates,
        ylabel=QuestEval (\%),
        xmin=0, xmax=34,
        ymin=40, ymax=50,
        xtick={4,8,16,32},
        ytick={41,42,43,44,45,46,47,48,49},
        legend style={nodes={scale=0.5, transform shape}, at={(0.50,0.95)}}
        ]
    \addplot[dashed,mark=*,myorange] plot coordinates {
        (4,44.74)
        (8,45.74)
        (16,46.65)
        (32,47.45)
    };
    \addlegendentry{Oracle [\texttt{QE}]}
    \addplot[smooth,mark=*,mygreen] plot coordinates {
        (4,43.36)
        (8,43.60)
        (16,43.83)
        (32,44.00)
     };
    \addlegendentry{EBR-ListMLE [\texttt{RL}]}
    \addplot[smooth,mark=*,myorange] plot coordinates {
        (4,43.78)
        (8,44.27)
        (16,44.74)
        (32,45.14)
    };
    \addlegendentry{EBR-ListMLE [\texttt{QE}]}
    \addplot[smooth,mark=*,myred] plot coordinates {
        (4,43.46)
        (8,43.78)
        (16,44.04)
        (32,44.25)
    };
    \addlegendentry{EBR-ListMLE [\texttt{Rel}]}
    \addplot[smooth,mark=*,myblue] plot coordinates {
        (4,43.46)
        (8,43.79)
        (16,44.05)
        (32,44.27)
    };
    \addlegendentry{EBR-ListMLE [\texttt{Cons}+\texttt{Rel}]}
    \addplot[smooth,mark=*,chocolate] plot coordinates {
        (4,43.17)
        (8,43.28)
        (16,43.32)
        (32,43.32)
    };
    \addlegendentry{BART}
    \end{axis}
\end{tikzpicture}
\end{subfigure}
\begin{subfigure}[b]{0.35\textwidth}
\centering
\begin{tikzpicture}
    \begin{axis}[
        xlabel=\# candidates,
        ylabel=$\textnormal{CTC}_\textnormal{consistency}$ (\%),
        xmin=0, xmax=34,
        ymin=90, ymax=100,
        xtick={4,8,16,32},
        ytick={91,92,93,94,95,96,97,98,99},
        legend style={nodes={scale=0.5, transform shape}, at={(0.50,0.325)}}
        ]
    \addplot[dashed,mark=*,myblue] plot coordinates {
        (4,95.44)
        (8,96.69)
        (16,97.54)
        (32,98.11)
    };
    \addlegendentry{Oracle [\texttt{Cons}]}
    \addplot[smooth,mark=*,mygreen] plot coordinates {
        (4,94.25)
        (8,95.01)
        (16,95.45)
        (32,95.70)
     };
    \addlegendentry{EBR-ListMLE [\texttt{RL}]}
    \addplot[smooth,mark=*,myorange] plot coordinates {
        (4,94.71)
        (8,95.71)
        (16,96.36)
        (32,96.78)
    };
    \addlegendentry{EBR-ListMLE [\texttt{QE}]}
    \addplot[smooth,mark=*,myred] plot coordinates {
        (4,94.85)
        (8,95.93)
        (16,96.66)
        (32,97.13)
    };
    \addlegendentry{EBR-ListMLE [\texttt{Rel}]}
    \addplot[smooth,mark=*,myblue] plot coordinates {
        (4,95.02)
        (8,96.15)
        (16,96.91)
        (32,97.41)
    };
    \addlegendentry{EBR-ListMLE [\texttt{Cons}+\texttt{Rel}]}
    \addplot[smooth,mark=*,chocolate] plot coordinates {
        (4,94.23)
        (8,95.01)
        (16,95.47)
        (32,95.77)
    };
    \addlegendentry{BART}
    \end{axis}
\end{tikzpicture}
\end{subfigure}
\;\hspace{2.5cm}
\begin{subfigure}[b]{0.35\textwidth}
\centering
\begin{tikzpicture}
    \begin{axis}[
        xlabel=\# candidates,
        ylabel=$\textnormal{CTC}_\textnormal{relevance}$ (\%),
        xmin=0, xmax=34,
        ymin=61, ymax=71,
        xtick={4,8,16,32},
        ytick={62,63,64,65,66,67,68,69,70},
        legend style={nodes={scale=0.5, transform shape}, at={(0.50,0.95)}}
        ]
    \addplot[dashed,mark=*,myred] plot coordinates {
        (4,63.53)
        (8,65.35)
        (16,66.99)
        (32,68.43)
    };
    \addlegendentry{Oracle [\texttt{Rel}]}
    \addplot[smooth,mark=*,mygreen] plot coordinates {
        (4,61.53)
        (8,62.16)
        (16,62.50)
        (32,62.84)
     };
    \addlegendentry{EBR-ListMLE [\texttt{RL}]}
    \addplot[smooth,mark=*,myorange] plot coordinates {
        (4,61.77)
        (8,62.48)
        (16,62.97)
        (32,63.27)
    };
    \addlegendentry{EBR-ListMLE [\texttt{QE}]}
    \addplot[smooth,mark=*,myred] plot coordinates {
        (4,62.41)
        (8,63.40)
        (16,64.14)
        (32,64.76)
    };
    \addlegendentry{EBR-ListMLE [\texttt{Rel}]}
    \addplot[smooth,mark=*,myblue] plot coordinates {
        (4,62.37)
        (8,63.32)
        (16,64.06)
        (32,64.64)
    };
    \addlegendentry{EBR-ListMLE [\texttt{Cons}+\texttt{Rel}]}
    \addplot[smooth,mark=*,chocolate] plot coordinates {
        (4,61.26)
        (8,61.75)
        (16,62.03)
        (32,62.14)
    };
    \addlegendentry{BART}
    \end{axis}
\end{tikzpicture}
\end{subfigure}
\caption{Performance of the models on the CNN/DailyMail dataset according to the indicated metrics for different numbers of candidate summaries. (\texttt{RL}: ROUGE-L, \texttt{QE}: QuestEval, \texttt{Cons}: $\textnormal{CTC}_\textnormal{consistency}$, \texttt{Rel}: $\textnormal{CTC}_\textnormal{relevance}$)}
\label{fig:num_candidates}
\end{figure*}

Figure~\ref{fig:num_candidates} shows the effect of varying the number of candidate summaries on the performance of our EBR models and BART baseline. The candidates were obtained using beam search with the number of beams equal to the number of candidates. The figure also shows the performance of the perfect re-ranker (Oracle), which defines the upper bound on the performance of the EBR.

Increasing the number of candidates leads to improvements in the performance of the EBR model when evaluated with the same metric it was trained to maximize. However, for ROUGE-L, these improvements are only marginal. Moreover, the performance gap between the Oracle and EBR tends to increase as well, especially in the reference-dependent metrics (ROUGE-L and $\textnormal{CTC}_\textnormal{relevance}$). The BART baseline also benefits from having larger beam sizes according to all metrics except ROUGE-L. Nonetheless, BART performs consistently worse than our EBR models according to all the metrics. Interestingly, increasing the number of candidates degrades ROUGE-L scores for all the models, except for EBR trained using this metric as the target.

\section{Human evaluation: further details}

\subsection{Evaluation form interface}
\label{sec:appendix_eval_form}

\begin{figure}
    \centering
    \includegraphics[max size={\textwidth}{\textheight}]{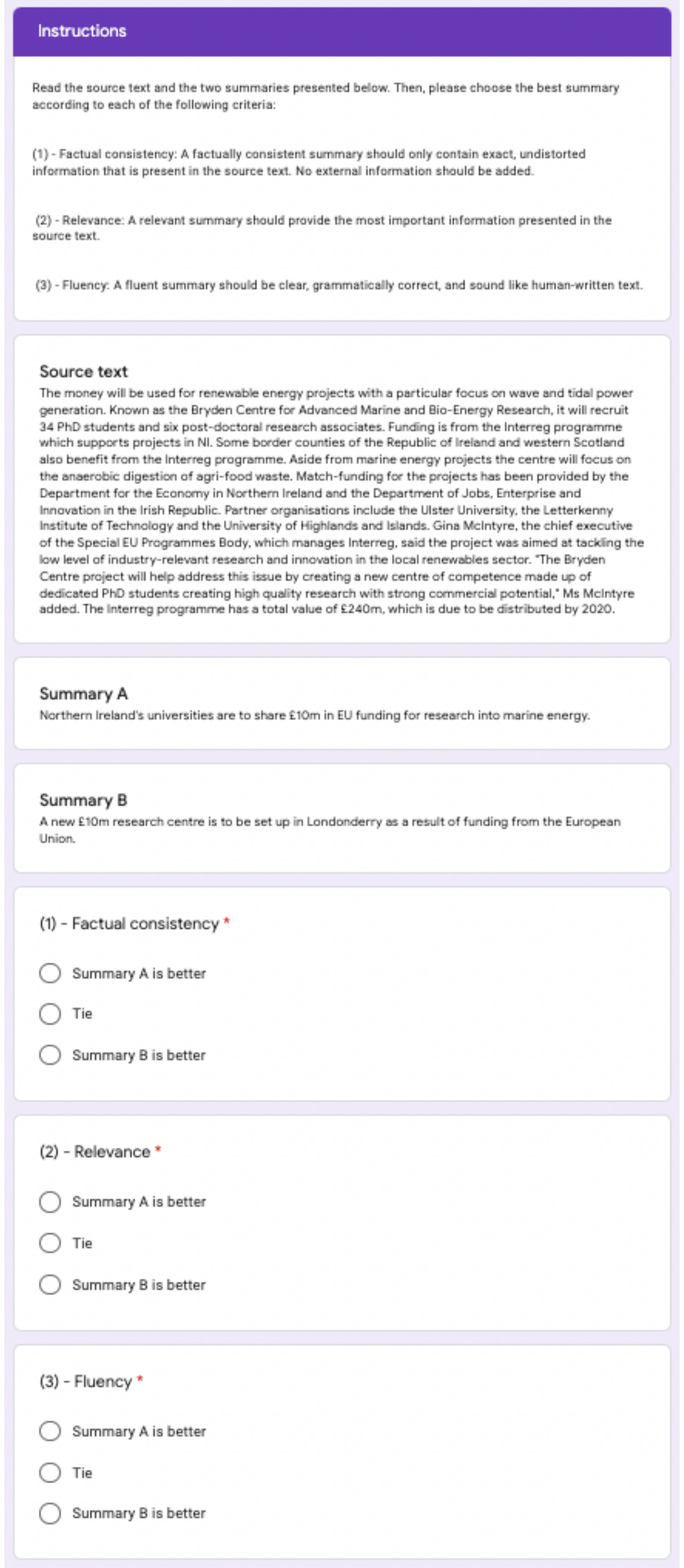}
    \caption{Evaluation form}
    \label{fig:eval_form}
\end{figure}

The human evaluation form was built using the Google Forms platform. Figure~\ref{fig:eval_form} presents a screenshot of the user interface. As we can observe, the interface was divided into seven sections. The first one provides instructions to the user and a brief definition of each of the three evaluation criteria: “(1) - Factual consistency: A factually consistent summary should only contain exact, undistorted information that is present in the source text. No external information should be added.”; “(2) - Relevance: A relevant summary should provide the most important information presented in the source text.”; “(3) - Fluency: A fluent summary should be clear, grammatically correct, and sound like human-written text.”. The three subsequent sections present the source text followed by the two anonymized summaries. Finally, the last three sections contain the multiple choice questions for each of the evaluation criteria. This seven-section pattern repeats itself for all pairwise comparisons in the evaluation form.

\subsection{Detected factual inconsistencies}
\label{sec:appendix_factual_inconsistencies}

In Table~\ref{tab:examples_bart_ebr} we show a few documents together with the summaries obtained from the baseline BART obtained with the usual beam search and the summaries chosen by the EBR model. Table~\ref{tab:examples_cliff_ebr} shows the examples from XSum used in the human evaluation questionnaire where the two judges agreed that the CLIFF summary was better than the EBR summary, regarding factual consistency. In two of the three examples, the $\textnormal{CTC}_\textnormal{consistency}$ metric wrongly assigns a larger score to the EBR summary than to the CLIFF summary. Interestingly, though, the EBR model would prefer the CLIFF summary over the BART summary in two of the three cases.

\begin{table*}
\begin{scriptsize}
\begin{tabular}{p{1cm}|p{11cm}cc}
       & Text                                                                                                                                                                                                                                                                                                                                                                                                                                                                                                                                                                                                                                                                                                                                                                                                                                                                                                                                                                                                                                                                                                                                                                                                                                                                                                                                                                                                                                                                                                                                                                                                                                                                                                                                                                                                                                                                                                                                                                                                                                                                                                                                                                                                                                                                                                                                                                                                                                                                                                                                                                                                                                                                                                                                                                                                                                                                                                                                                                                                                                                                                                                                                                                                                                                                                                                                                                                                                                                                                                                                                                                                                                                                                                                                                                                                                                                                                                                                                                                                                                                                      & $\texttt{Cons}$ & $E$ \\
\end{tabular}
\end{scriptsize}
\caption{Examples where the judges agreed that one of the summaries was better than the other on the factual consistency dimension. Consistent and inconsistent segments are highlighted in \textcolor{green!50!black}{green} and \textcolor{red!80!black}{red}, respectively. Columns $\texttt{Cons}$ and $E$ show the $\textnormal{CTC}_\textnormal{consistency}$ (in \%) and the energy score (output of the EBR model) on each of the summaries, respectively. (Remember that for $E$ lower is better.)}
\label{tab:examples_bart_ebr}
\end{table*}

\begin{table*}
\begin{scriptsize}
\begin{tabular}{p{1cm}|p{11cm}cc}
       & Text                                                                                                                                                                                                                                                                                                                                                                                                                                                                                                                                                                                                                                                                                                                                                                                                                                                                                                                                                                                                                                                                                                                                                                                                                                                                                                                                                                                                                                                                                                                                                                                                                                                                                                                                                                                                                                                                                                                                                                                                                                                                                                                                                                                                                                                                                                                                                                                                                                                                                                                                                                                                                                                                                                                                                                                                                                                                                                                                                                                                                                                                                                                                                                                                                                                                                                                                                                                                                                                                                                                                                                                                                                                                                                                                                                                                                                                                                                                                                                                                                                                                      & $\texttt{Cons}$ & $E$ \\
\end{tabular}
\end{scriptsize}
\caption{Examples from XSum where the two judges agreed that CLIFF was better than EBR on the factual consistency dimension. Consistent and inconsistent segments are highlighted in \textcolor{green!50!black}{green} and \textcolor{red!80!black}{red}, respectively. Columns $\texttt{Cons}$ and $E$ show the $\textnormal{CTC}_\textnormal{consistency}$ (in \%) and the energy score (output of the EBR model) on each of the summaries, respectively. (Remember that for $E$ lower is better.)}
\label{tab:examples_cliff_ebr}
\end{table*}

\end{document}